\documentclass[letterpaper, 10 pt, conference]{ieeeconf}

\ifdefined\pdfminorversion
\fi

\IEEEoverridecommandlockouts
\usepackage{graphicx}
\usepackage{booktabs}
\usepackage{nicematrix}
\usepackage{multirow}
\usepackage{amsmath,amssymb}
\usepackage{newtxtext,newtxmath}
\usepackage[table]{xcolor}

\usepackage{cite}

\makeatletter
\let\NAT@parse\undefined
\makeatother

\usepackage{hyperref}
\hypersetup{
    colorlinks=true,
    citecolor=green!50!black,
    linkcolor=red,
    urlcolor=blue      
}

\usepackage[all]{hypcap}

\usepackage{cuted}
\usepackage{capt-of}

\newcommand{\dataset}{RiverVLN}
\newcommand{\policy}{PGT-NAV}

\title{\LARGE \bfseries
RiverVLN: Phase-Grounded Temporal Vision--Language Navigation for Unmanned Surface Vehicles}

\author{Jieling Wu, Yuehao Huang, Jiajun Lv, Tao Huang, Yong Liu, Weiwei Liu}

\begin{document}
\bstctlcite{IEEEexample:BSTcontrol}
\maketitle
\thispagestyle{empty}
\pagestyle{empty}

\begin{strip}
    \centering
    \includegraphics[width=0.98\textwidth]{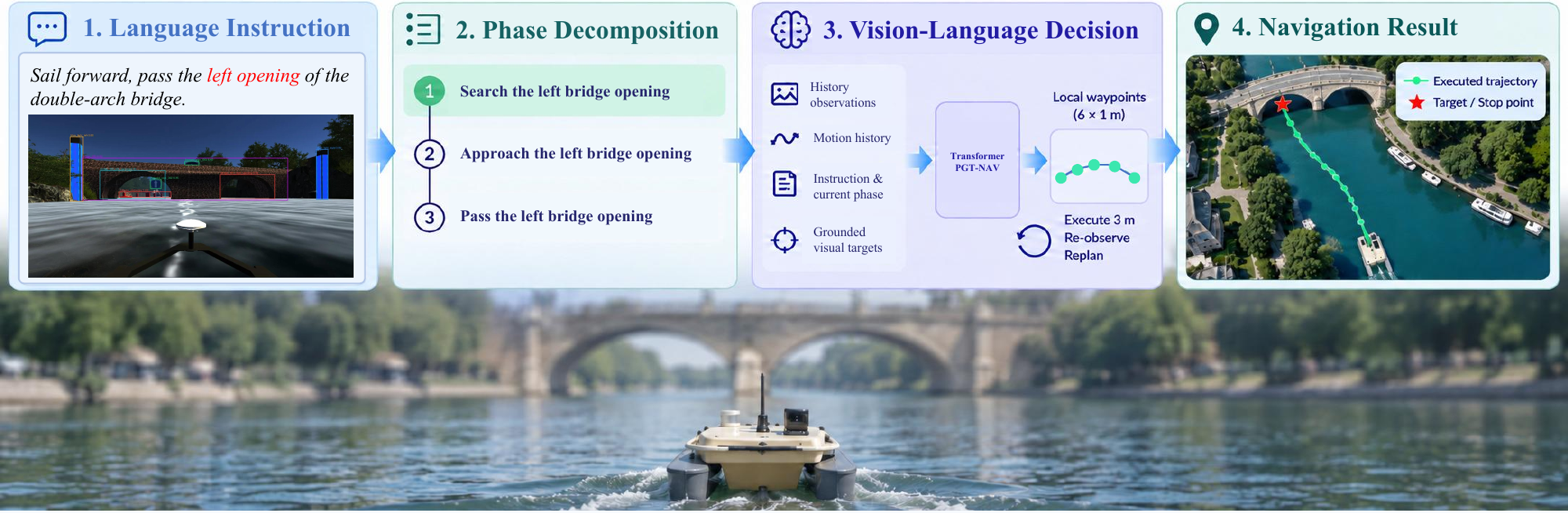}
    \captionof{figure}{\textbf{Overview of RiverVLN.} A natural-language instruction is decomposed into ordered semantic phases, and phase-grounded visual--motion context is used to predict short-horizon waypoints for closed-loop USV navigation.}
    \label{fig:overview}
\end{strip}

\begin{abstract}

Vision--language navigation (VLN) has largely been developed for indoor and terrestrial robots, where language can often be treated as a static goal and motion is approximated by discrete or near-instantaneous actions. These assumptions break down for unmanned surface vehicles (USVs): river navigation requires continuous motion under inertia and limited maneuverability, while long-horizon instructions must be executed through sparse and visually ambiguous maritime landmarks.
We introduce \dataset{}, to our knowledge the first benchmark designed for long-horizon USV VLN under continuous riverine motion, and \policy{}, a phase-grounded temporal navigation framework for USVs. Rather than directly mapping an entire instruction to motion, \policy{} converts it into an ordered sequence of visually verifiable semantic phases and maintains the active phase online through grounded visual and motion evidence. This explicit semantic progress state is fused with visual--motion history and phase-specific grounding to predict six local $\mathrm{SE}(2)$ pose increments. The resulting trajectory is executed in a predict--execute--re-observe loop, where the vessel executes toward $W_3$, updates phase and grounding, and replans through a map-based safety layer.
Experiments show that \policy{} substantially reduces recursive position and heading drift relative to GNM-style and ViNT-style baselines and achieves an average success rate of 0.79 in Unity--ROS closed-loop navigation. Unseen bridge-opening trials and real-world USV experiments further demonstrate that the phase-grounded representation transfers from controlled evaluation to physical USV deployment.

\end{abstract}

\section{INTRODUCTION}
Unmanned surface vehicles (USVs) are increasingly deployed for environmental monitoring, waterway inspection, and infrastructure-related operations~\cite{hu2022colregs_review,ozturk2022maritime_review}.
As riverine applications expand, natural-language instructions provide an intuitive interface for specifying navigation tasks without requiring operators to define low-level waypoints or control commands.
Such tasks often require considerably more than reaching a destination: a vessel may be instructed to follow a channel, select one bridge opening among visually similar alternatives, pass semantic landmarks in a specified order, turn relative to a directional cue, and finally stop near a designated target. Successfully executing these instructions therefore requires reasoning about temporal order, object attributes, spatial relations, and terminal conditions, rather than simply navigating toward a fixed goal.

Vision--Language Navigation (VLN) studies how an embodied agent grounds natural-language instructions in first-person visual observations and translates them into navigation decisions. It has been extensively studied in indoor and terrestrial embodied environments~\cite{krantz2020vlnce,hong2021vlnbert,chen2021hamt,zhang2026sparse} and has also been extended to aerial platforms~\cite{liu2023aerialvln}. Recent embodied vision--language models have further improved instruction following through richer multimodal representations~\cite{zheng2024navillm,zhang2024navid,huang2025cogddn,huang2026wnm,li2026gn0}. These advances provide a strong foundation for language-conditioned navigation, but they do not directly address how semantic instruction progress interacts with the continuous dynamics of a surface vessel.

Extending VLN to river navigation, however, introduces challenges arising from both the characteristics of river environments and the dynamics of USVs.
First, river environments often contain sparse semantic landmarks and repetitive visual structure, making visually similar targets difficult to distinguish and a single observation insufficient to determine whether an instruction-relevant landmark is still being approached or has already been passed.
Navigation therefore requires temporal evidence to infer and maintain the current semantic progress of the instruction.
Second, USVs exhibit strongly non-instantaneous motion dynamics, including pronounced inertia, lateral drift, delayed stopping, and a finite turning radius. Consequently, an incorrect semantic transition cannot always be corrected by an immediate change of action. For example, prematurely switching from an approach phase to a pass phase, or initiating a terminal maneuver too late, may already place the vessel on an undesirable trajectory.
Riverine VLN therefore requires semantic progress estimation and continuous motion prediction to be considered jointly, rather than treating instruction grounding and short-horizon control as largely independent problems.

Studying these coupled challenges requires a benchmark that jointly captures long-horizon language instructions, temporally evolving semantic progress, visual observations, and continuous vessel motion.
To this end, we introduce \dataset{}, a Unity--ROS benchmark for long-horizon VLN in continuous river environments. \dataset{} aligns natural-language instructions with visual--motion observations, metric trajectories, phase-level task progress, and instruction-relevant maritime grounding annotations. To our knowledge, it is the first benchmark designed for long-horizon USV VLN under continuous riverine motion.
Building on \dataset{}, we propose \textbf{\policy{}}, a \textbf{P}hase-\textbf{G}rounded \textbf{T}emporal \textbf{Nav}igation framework. Rather than predicting motion directly from the current observation and the full instruction, \policy{} maintains an explicit estimate of the active semantic phase, grounds the phase-relevant target in the current observation, and jointly reasons over visual--motion history to predict a short-horizon metric trajectory.

Our contributions are summarized as follows:
\begin{itemize}
    \item We introduce \dataset{}, a continuous-river VLN benchmark for USVs that pairs long-horizon navigation instructions with visual--motion observations, metric trajectories, phase-level task progress, and phase-aligned maritime grounding annotations.
    
    \item We propose \policy{}, a phase-grounded temporal framework that jointly models visual--motion history, explicit semantic progress, and phase-specific grounding to predict short-horizon metric trajectories for continuous USV navigation.
    
    \item We evaluate \policy{} through offline trajectory prediction, Unity--ROS closed-loop navigation, and real-world USV experiments, demonstrating improved closed-loop performance in simulation and the feasibility of deploying the proposed framework on a real USV.
\end{itemize}

The dataset, code, and evaluation tools will be publicly released at
\url{https://anonymous.4open.science/r/PGT-NAV-D349}.

\section{RELATED WORK}

\subsection{Autonomous USV Navigation and Motion Planning}

Autonomous USV navigation commonly combines global path planning, collision avoidance, and local control, with COLREGs-aware risk assessment and replanning central to operation in constrained waterways \cite{hu2022colregs_review,ozturk2022maritime_review}. Recent systems further account for current fields, energy consumption, and dynamic obstacles through hierarchical planning and optimization \cite{zhao2023hierarchical,hao2023current}. Deterministic methods based on A*, dynamic-window search, and modified artificial potential fields remain attractive because their geometric constraints are interpretable and can be coupled directly with vessel kinematics and safety margins \cite{liu2023colregs_apf,hu2023astar_dwa}.
Learning-based approaches, including deep reinforcement learning, have also been explored for local collision avoidance and COLREGs-compliant decisions \cite{wen2023drl}. These methods are effective when the navigation objective is represented by waypoints, coordinates, occupancy maps, or explicit rules, but they generally do not model long natural-language instructions containing ordered semantic goals, object attributes, relative directions, and stopping conditions. Our framework therefore does not replace deterministic planning: it retains map-based planning as the safety execution layer and places a language-conditioned semantic navigation layer above it to determine the task-relevant local intent.

\subsection{VLN in Continuous Environments}

Vision--language navigation was initially studied on discrete indoor navigation graphs, where agents selected among graph-connected actions. VLN-CE extended instruction following to continuous 3-D environments, exposing the agent to visual interpretation, traversability, and low-level motion execution \cite{krantz2020vlnce}.
Recurrent cross-modal models and history-aware Transformers subsequently improved progress reasoning \cite{hong2021vlnbert,chen2021hamt}, while waypoint prediction and discrete-to-continuous transfer improved geometric executability \cite{krantz2021waypoint,hong2022bridging}. Long-horizon VLN further motivated explicit memory and structured spatial representations: DUET and cross-modal map learning combine local visual reasoning with graph-based or bird's-eye representations \cite{chen2022duet,georgakis2022crossmodal}, milestone-based navigation decomposes long instructions into ordered intermediate goals \cite{song2022milestones}, and ScaleVLN and GOAT-Bench emphasize data scale, multimodal goals, and long-term memory \cite{wang2023scalevln,khanna2024goat}. Despite these advances, most continuous VLN methods assume terrestrial embodiments whose turning and stopping dynamics differ substantially from those of USVs. River scenes additionally contain sparse maritime landmarks and repetitive structures such as shorelines and bridge openings, making task progress more ambiguous. Our method is related to milestone- and progress-aware VLN, but represents progress explicitly as a semantic phase tied to a grounded maritime target and predicts a short-horizon metric $\mathrm{SE}(2)$ trajectory that is repeatedly replanned during execution rather than a sequence of discrete actions.

\subsection{Foundation Models for Language-Guided Navigation}

Recent foundation-model-based navigation systems combine pretrained language, vision, and navigation models. LM-Nav couples large language and vision models with a navigation policy for outdoor landmark following~\cite{shah2023lmnav}. GNM~\cite{shah2023gnm} and ViNT~\cite{shah2023vint} learn transferable visual navigation from heterogeneous robot data, with ViNT using a Transformer to fuse temporal observations and navigation goals. NoMaD extends this family with a diffusion policy for goal reaching and exploration~\cite{sridhar2024nomad}. More recent systems extend beyond image-goal conditioning: NaviLLM, NaVid, and Uni-NaVid use language- or video-conditioned prediction~\cite{zheng2024navillm,zhang2024navid,zhang2025uninavid}, VLFM couples vision--language semantics with frontier planning~\cite{yokoyama2024vlfm}, and CA-Nav reasons about sub-instruction completion in continuous environments~\cite{chen2025canav}. In contrast, we focus on long-horizon river navigation under USV dynamics and sparse maritime landmarks. We invoke the language model only for initial phase decomposition, while online progress is updated from grounded visual and motion evidence, linking phase-specific grounding and temporal progress tracking to metric waypoint prediction.

\section{METHOD}

\begin{figure}[!t]
    \centering
    \includegraphics[width=0.95\columnwidth]{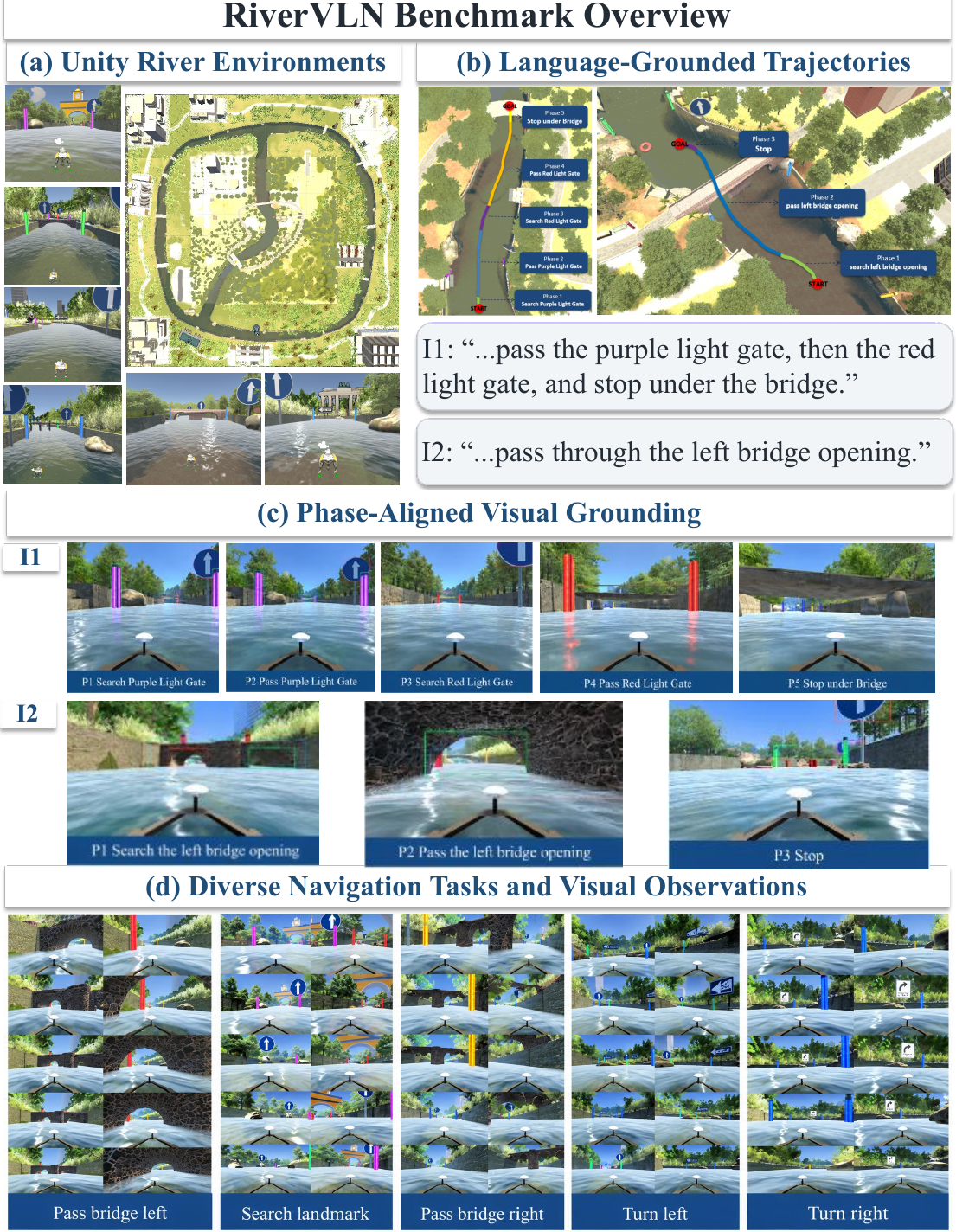}
    \caption{\textbf{Overview of the \dataset{} benchmark.} 
    (a) Unity river environments and global layouts.
    (b) Language-grounded trajectories aligned with long-horizon instructions. 
    (c) Phase-aligned visual grounding annotations.
    (d) Representative tasks including landmark search, bridge selection, and
    directional navigation.}
    \label{fig:dataset}
\vspace{-2mm}
\end{figure}

\begin{figure*}[t]
    \centering
    \includegraphics[width=1\textwidth]{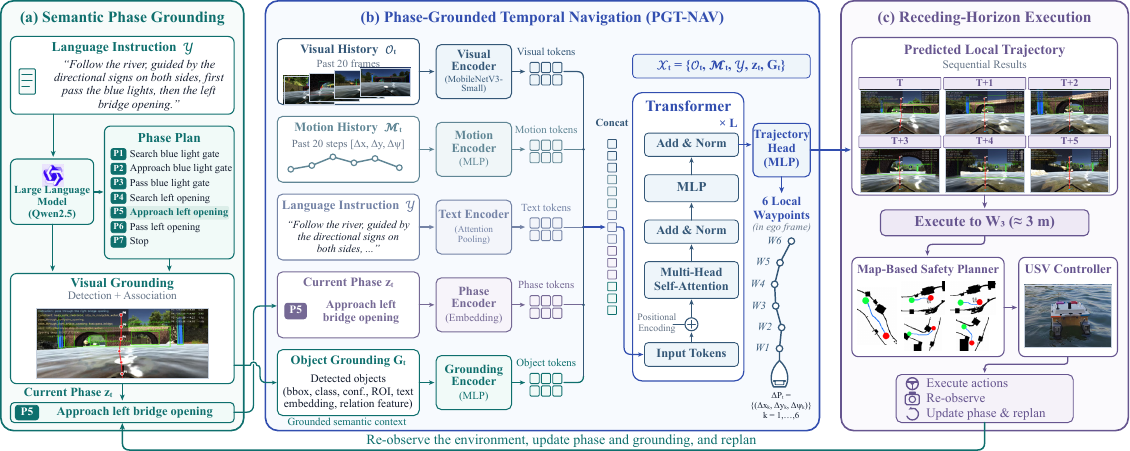}
    \caption{\textbf{Overview of \policy{}. }
    (a) Long-horizon instructions are decomposed into ordered semantic phases and associated with visually grounded targets. 
    (b) Visual--motion history, language, the current phase, and phase-specific grounding cues are fused by a Transformer to predict six local $\mathrm{SE}(2)$ waypoints. 
    (c) The controller executes toward $W_3$, re-observes the environment, updates phase and grounding, and replans in a receding-horizon loop.}
    \label{fig:framework}
\end{figure*}

Given a natural-language navigation instruction $I$, the USV must produce an executable continuous trajectory while respecting the instruction's semantic order, spatial constraints, and stopping condition. We formulate this as phase-conditioned short-horizon trajectory prediction rather than direct discrete action generation. The overall framework is shown in Fig.~\ref{fig:framework}.

\subsection{RiverVLN Dataset and Task Formulation}

Constructing a VLN benchmark for USVs requires addressing
three data-level challenges. First, sparse maritime landmarks
and language-specified spatial relations make it difficult
to associate navigation instructions with the intended
visual targets. We address the lack of explicit
language-to-target supervision by annotating object
identities, bounding boxes, and spatial relations, thereby
linking each instruction to its corresponding maritime
landmarks. Second, vessel inertia and limited maneuverability
require navigation decisions to account for continuous
motion rather than isolated observations or discrete
actions. We address the lack of motion-aware supervision
by recording continuous pose histories under constrained
vessel control and deriving metric displacement, heading,
and future waypoint labels. Third, long-horizon instructions
contain ordered sub-tasks and stopping conditions that
cannot be represented by a terminal goal alone. We address
the lack of explicit semantic-progress supervision by
aligning instructions with ordered phase labels, active
targets, and corresponding trajectory segments, including
terminal stopping phases.

As shown in Fig.~\ref{fig:dataset}, \dataset{} is constructed
in Unity--ROS and covers river routes containing bridges
with multiple openings, directional signs, colored light
gates, shoreline landmarks, and obstacles. Object identities,
grounding boxes, and spatial-relation annotations establish
the correspondence between language-specified targets and
visual observations, including individual bridge openings.
A forward-facing RGB camera captures $1280\times720$ images
with a $60^\circ$ vertical and approximately $91.5^\circ$
horizontal field of view; images are resized to
$512\times288$ for network input. To retain metric motion
information, navigation samples are recorded at approximately
$1\,\mathrm{m}$ intervals of accumulated travel rather than
at fixed time intervals. Reference trajectories are collected
using a $20\,\mathrm{Hz}$ ROS controller with a maximum
commanded speed of $2.0\,\mathrm{m/s}$, a maximum commanded
yaw rate of $1.0\,\mathrm{rad/s}$, and a $2.0\,\mathrm{m}$
minimum turning-radius constraint during normal forward
tracking. Synchronized RGB and pose records preserve the
vessel's motion history and yield body-frame displacement
and heading increments, providing continuous
$\mathrm{SE}(2)$ trajectory supervision. Each route is
further paired with natural-language instructions and
ordered phase plans. Phase labels, active targets, grounding
boxes, spatial relations, and trajectory segments are aligned
along the route, identifying the objective of each segment
and explicitly representing terminal stopping behavior.

A navigation episode is represented as
\begin{equation}
\mathcal{D}^{(n)}=
\left(
I^{(n)},\mathcal{Z}^{(n)},
\{o_t,p_t,m_t,z_t,G_t\}_{t=1}^{T_n}
\right),
\label{eq:dataset_episode}
\end{equation}
where $I^{(n)}$ and $\mathcal{Z}^{(n)}$ denote the instruction
and ordered phase plan, while $o_t$, $p_t$, $m_t$, $z_t$,
and $G_t$ represent the RGB observation, vessel pose,
body-frame motion increment, phase label, and phase-specific
grounding, respectively. At each decision step, the dataset
constructs the PGT-NAV input
\begin{equation}
\mathcal{X}_t=
(\mathcal{O}_t,\mathcal{M}_t,I,z_t,G_t),
\label{eq:dataset_input}
\end{equation}
where $\mathcal{O}_t$ and $\mathcal{M}_t$ contain the latest
$H=20$ visual and motion observations, with missing history
masked. Six future local $\mathrm{SE}(2)$ waypoint increments
are derived from reference poses as prediction targets.
Consequently, each sample jointly provides language-grounded
visual targets, continuous motion context, and phase-aligned
trajectory supervision, enabling PGT-NAV to learn local
navigation behavior conditioned on the current semantic
objective. Table~\ref{tab:dataset_stats} summarizes the
Train, Val, and Test-Unseen splits, comprising 260 routes,
1,040 instructions, 108,336 samples, 6,134 phase instances,
and 103,472 grounding boxes.

\begin{table}[!t]
\centering
\caption{Statistics of the \dataset{} benchmark}
\label{tab:dataset_stats}
\setlength{\tabcolsep}{3pt}
\small
\textbf{Annotation Statistics}\par\vspace{2pt}
\begin{NiceTabular*}{\columnwidth}{@{\extracolsep{\fill}}l c l c}
\toprule
\textbf{Annotation} & \textbf{Count} &
\textbf{Annotation} & \textbf{Count} \\
\midrule
Phase Categories  & 7  & Phase Instances & 6,134 \\
Object Categories & 16 & Grounding Boxes & 103,472 \\
\bottomrule
\end{NiceTabular*}
\par\vspace{6pt}
\textbf{Dataset Statistics}\par\vspace{2pt}
\resizebox{\columnwidth}{!}{
\begin{NiceTabular}{lrrrrrr}
\toprule
\textbf{Split} & \textbf{Routes} & \textbf{Samples} & \textbf{RGB} &
\textbf{Instr.} & \textbf{Len.} & \textbf{Phases} \\
\midrule
Train       & 201 & 88,388 & 22,016 & 804 & 108.43 & 5.94 \\
Val         & 36  & 13,172 & 3,277  & 144 & 89.97  & 5.38 \\
Test-Unseen & 23  & 6,776  & 1,688  & 92  & 72.22  & 6.35 \\
\midrule
Total       & 260 & 108,336 & 26,981 & 1,040 & 102.67 & 5.90 \\
\bottomrule
\end{NiceTabular}}
\par
\end{table}

\subsection{Semantic Phase Grounding}

Long-horizon river navigation requires the USV to track
which sub-task is currently active rather than treating
the instruction as a static destination. At the beginning
of each episode, Qwen2.5-7B-Instruct is invoked once to
decompose instruction $I$ into an ordered phase plan.
Its output is restricted to seven phases:
\emph{search}, \emph{approach}, \emph{maneuver},
\emph{pass}, \emph{stop}, \emph{stop-before-target},
and \emph{stop-near-target}. The decomposition preserves
directional, color, landmark, and stopping constraints
without introducing new targets. For example, ``pass
through the left bridge opening'' yields
\emph{search} $\rightarrow$ \emph{approach}
$\rightarrow$ \emph{pass} for the specified opening.
The active phase provides a local semantic objective,
preventing intermediate targets from being confused
with the terminal destination.

During execution, detections are associated with the
active target using object class and language-specified attributes, and only phase-relevant detections form $G_t$.
Rather than switching phases according to a fixed
schedule or a single image, a deterministic state
machine evaluates recent visual and motion evidence
$\mathcal{E}_t$, including target visibility,
normalized box area, horizontal image position,
and vessel motion:
\begin{equation}
z_{t+1}=
\begin{cases}
\operatorname{next}(z_t),
& C_{z_t}(\mathcal{E}_t)=1,\\
z_t, & \text{otherwise},
\end{cases}
\label{eq:phase_update}
\end{equation}
where $C_{z_t}$ is a phase-specific transition
predicate. In particular, target disappearance alone
cannot confirm passage because it may result from
viewpoint changes or detection failures. Passage
completion therefore also requires traveled-distance
evidence, while directional maneuvers incorporate
accumulated heading change.

Search-to-approach transitions use persistent target visibility and scale, whereas approach-to-pass transitions additionally consider image position and vessel motion.
To suppress detection noise, ordinary transitions
require at least five positive observations among
the latest eight and a minimum phase age of 15
observations. Ordering guards prevent future targets
from advancing the current phase prematurely.
The manager thus updates semantic progress online
without repeated LLM calls. Ground-truth phase labels
are used for training supervision and evaluation
metrics only, never for online phase selection.

\subsection{Phase-Grounded Temporal Trajectory Prediction}

Due to vessel inertia and limited maneuverability,
identical visual observations may require different
trajectories depending on recent motion. Inspired by
ViNT \cite{shah2023vint}, we condition local trajectory
prediction on visual and motion histories, instruction
$I$, active phase $z_t$, and phase-specific grounding
$G_t$. The phase specifies the immediate objective,
grounding identifies its visual target, and motion
history captures recent translation and rotation.
These modalities are fused by a Transformer:
\begin{equation}
\begin{aligned}
X_t={}&[E_v(\mathcal{O}_t),E_m(\mathcal{M}_t),
E_l(I),E_s(z_t),E_g(G_t)],\\
F_t={}&\mathrm{Transformer}(X_t),
\end{aligned}
\label{eq:fusion}
\end{equation}
where the encoders produce modality-specific tokens
and $\mathcal{O}_t,\mathcal{M}_t$ contain the latest
$H=20$ observations. This temporal conditioning
allows the policy to account for ongoing vessel
motion instead of relying on a single image.

The trajectory head predicts six body-frame
$\mathrm{SE}(2)$ increments:
\begin{equation}
\hat{\Delta P}_t=f_{\mathrm{traj}}(F_t)
=\{(\Delta x_k,\Delta y_k,\Delta\psi_k)\}_{k=1}^{6},
\label{eq:trajectory}
\end{equation}
with approximately $1\,\mathrm{m}$ of progress per
step. The increments represent a local trajectory independent of global-map coordinates,
while their accumulated waypoints describe the
short-horizon turning pattern. Unlike predicting
isolated positions or discrete steering actions,
this representation provides a sequence of
displacement and heading changes for subsequent
execution. Map-based obstacle handling remains
the responsibility of the downstream safety planner.

Training jointly supervises the predicted increments,
their accumulated waypoints, and the execution anchor
$W_3$. Position errors use Smooth-L1 ($\beta=1$),
and heading errors use the circular distance
$e_\psi=|\operatorname{atan2}(
\sin(\hat\psi-\psi),\cos(\hat\psi-\psi))|$.
The three terms use component weights $(1,3,12)$,
$(1,3,8)$, and $(1,4,16)$, respectively. The
overall objective is
\begin{equation}
\begin{aligned}
\mathcal{L}={}&\mathcal{L}_{\Delta}
+0.5\mathcal{L}_{W}
+2\mathcal{L}_{\mathrm{exec}}\\
&+\mathcal{L}_{\mathrm{straight}}
+0.25\mathcal{L}_{\mathrm{relation}}
+\mathcal{L}_{\mathrm{stop}},
\end{aligned}
\label{eq:training_objective}
\end{equation}
where the auxiliary terms regularize near-straight
motion, target alignment, and stopping behavior.
The execution-anchor loss emphasizes $W_3$, which
defines the execution target, aligning
training with the receding-horizon strategy.

\subsection{Receding-Horizon Closed-Loop Execution}

Executing the full predicted trajectory open loop
can accumulate motion errors before the USV receives
new observations. We therefore execute only part
of each prediction and replan from the updated
vessel state, allowing the trajectory to adapt
to both motion deviations and changes in the
active semantic objective.

At each decision step, the six predicted
$\mathrm{SE}(2)$ increments are converted into
local waypoints spanning approximately $6\,\mathrm{m}$.
The third waypoint $W_3$, approximately
$3\,\mathrm{m}$ ahead, serves as the execution
anchor. The predicted corridor and known map
are passed to a safety planner
\cite{zhao2023hierarchical,hu2023astar_dwa},
whose path is tracked by the low-level controller.
The policy thus provides language-conditioned
motion intentions, while planning and control
handle map-based obstacles and path tracking.

After approximately $3\,\mathrm{m}$ of execution,
the USV acquires new visual and motion observations,
updates its phase and target grounding, and predicts
a trajectory. The unexecuted waypoints from
the previous prediction are discarded. As shown
in Fig.~\ref{fig:framework}, this forms a
\emph{predict 6\,m $\rightarrow$ execute 3\,m
$\rightarrow$ re-observe $\rightarrow$ replan}
loop, continuously revising the trajectory
according to updated motion and semantic progress.
\section{EXPERIMENTS}
\label{sec:experiments}

We evaluate \policy{} at three complementary levels. Offline recursive rollout measures trajectory and heading stability under repeated predictions, ablations isolate the contributions of phase, temporal history, and grounding, and Unity--ROS closed-loop evaluation and real-world USV trials test whether these gains translate to executable navigation.

\begin{table*}[t]
\centering
\caption{Offline comparison and ablation on the \dataset{} validation split}
\label{tab:offline_all}
\small
\setlength{\tabcolsep}{5.5pt}
\renewcommand{\arraystretch}{1.02}
\begin{NiceTabular}{lccccc}
\CodeBefore
\rowcolor{blue!10}{7}
\Body
\toprule
\textbf{Method} &
\textbf{Teacher ADE (m)} &
\textbf{Anchor Yaw Err. ($^\circ$)} &
\textbf{Rec. ADE (m)} &
\textbf{Rec. FDE (m)} &
\textbf{Exec. Yaw Err. ($^\circ$)} \\
\midrule
GNM-style~\cite{shah2023gnm}
& 0.0696 & 7.8054 & 6.2387 & 6.5177 & 8.0644 \\
ViNT-style~\cite{shah2023vint}
& 0.0476 & 11.2238 & 9.8883 & 10.2982 & 11.7040 \\
\midrule
w/o Phase
& 0.1247 & 13.9324 & 13.3150 & 13.8202 & 14.4716 \\
w/o Temporal
& 0.1269 & 13.9265 & 13.7390 & 14.2165 & 14.4441 \\
w/o Grounding
& \textbf{0.0557} & 9.1439 & 6.7696 & 7.0825 & 9.5724 \\
\textbf{PGT-NAV}
& 0.0858 & \textbf{3.2770} & \textbf{2.8171}
& \textbf{2.9080} & \textbf{3.4283} \\
\bottomrule
\end{NiceTabular}
\end{table*}

\subsection{Experimental Setup and Protocol}

Train is used for optimization, Val for checkpoint selection and offline evaluation, and Test-Unseen only for generalization evaluation. In offline rollout, the first state is anchored to ground truth and each subsequent decision uses the predicted $W_3$ pose as the next anchor at approximately 3\,m intervals. Because the RGB stream remains on the recorded reference trajectory, these errors measure sensitivity to recursive pose drift rather than closed-loop navigation success. GNM-style and ViNT-style retain their visual-navigation principles but are adapted to the same history length, decision interval, and six-waypoint output representation; neither receives the semantic phase or phase-specific grounding available to \policy{}, and all methods are evaluated on the same Val trajectories and rollout anchors.

For Unity--ROS closed-loop evaluation, the high-level policy runs every 2\,s and the low-level controller at 20\,Hz, with maximum USV speed limited to 1.5\,m/s. The forward RGB camera publishes $1280\!\times\!720$ images at 10\,Hz, which are resized to $512\!\times\!288$ before inference. The policy uses 20 history frames, a frozen pretrained MobileNetV3-Small visual backbone, and a 3-layer Transformer with 256-dimensional embeddings and four attention heads. Training uses AdamW for up to 200 epochs with batch size 8, gradient accumulation 4, and learning rate and weight decay of $10^{-4}$; the checkpoint with the best validation performance is used for evaluation.

\subsection{Offline Prediction and Ablation}

We report Teacher ADE, Anchor Yaw Error, Recursive ADE/FDE, and Execution Yaw Error. The first measures prediction from a ground-truth anchor; the remaining metrics measure positional and heading drift under recursive rollout. Because the RGB stream remains on the recorded trajectory, recursive errors are not interpreted as closed-loop success.

Table~\ref{tab:offline_all} shows that GNM-style and ViNT-style achieve low teacher-forced error but accumulate substantially larger recursive drift. In contrast, \policy{} maintains lower recursive position and heading errors. Removing phase or temporal history raises Recursive ADE/FDE to roughly 13--14\,m, while removing grounding increases Recursive ADE/FDE to 6.770/7.083\,m despite a lower Teacher ADE. The gain therefore lies primarily in maintaining semantic-motion consistency across repeated decisions rather than in minimizing single-step regression error.

\begin{figure}[!t]
    \centering
    \includegraphics[width=0.84\linewidth]{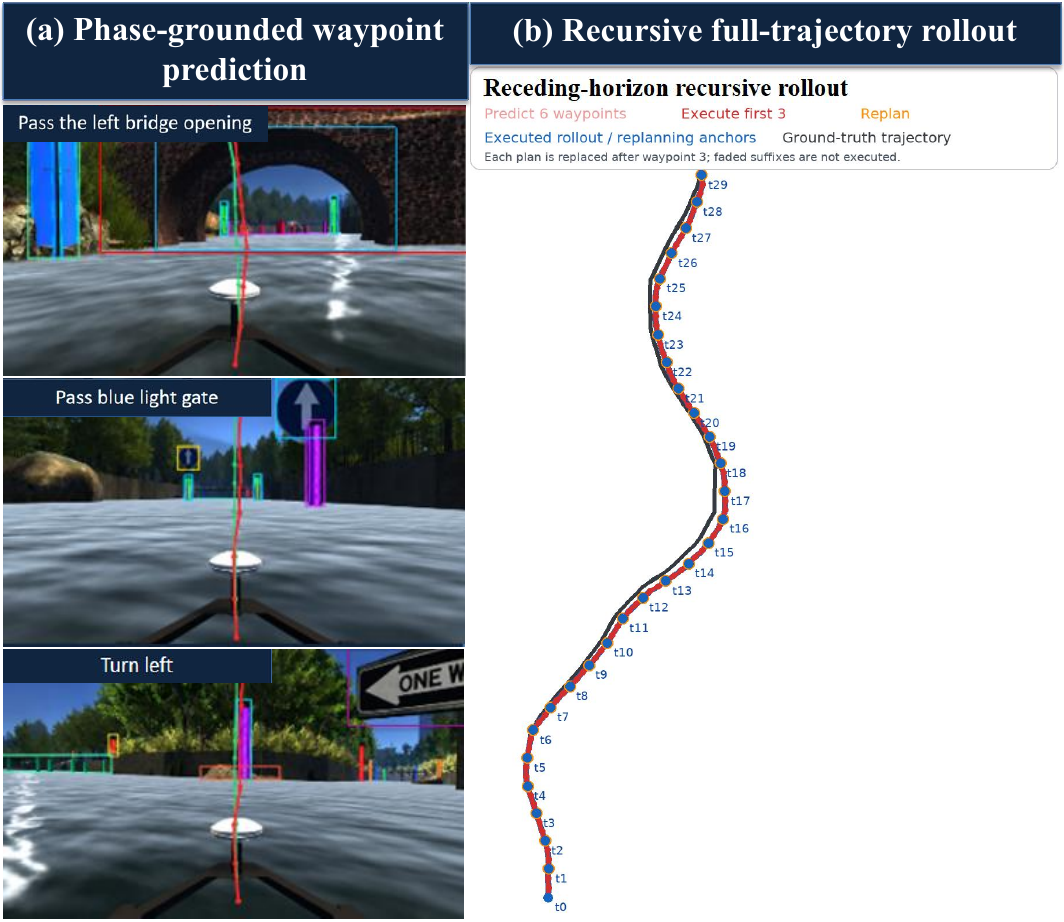}
    \caption{\textbf{Local waypoint prediction and recursive rollout.} The examples on the left show phase-conditioned local predictions; the global rollout on the right visualizes error propagation under repeated prediction and execution anchors.}
    \label{fig:prediction}
\end{figure}

Figure~\ref{fig:prediction} illustrates the effect: local predictions remain aligned with the intended bridge, light gate, or turn, whereas recursive rollout exposes accumulated state error. This motivates re-observation after each 3\,m execution segment.

\begin{figure*}[!t]
    \centering
    \includegraphics[width=0.95\textwidth]{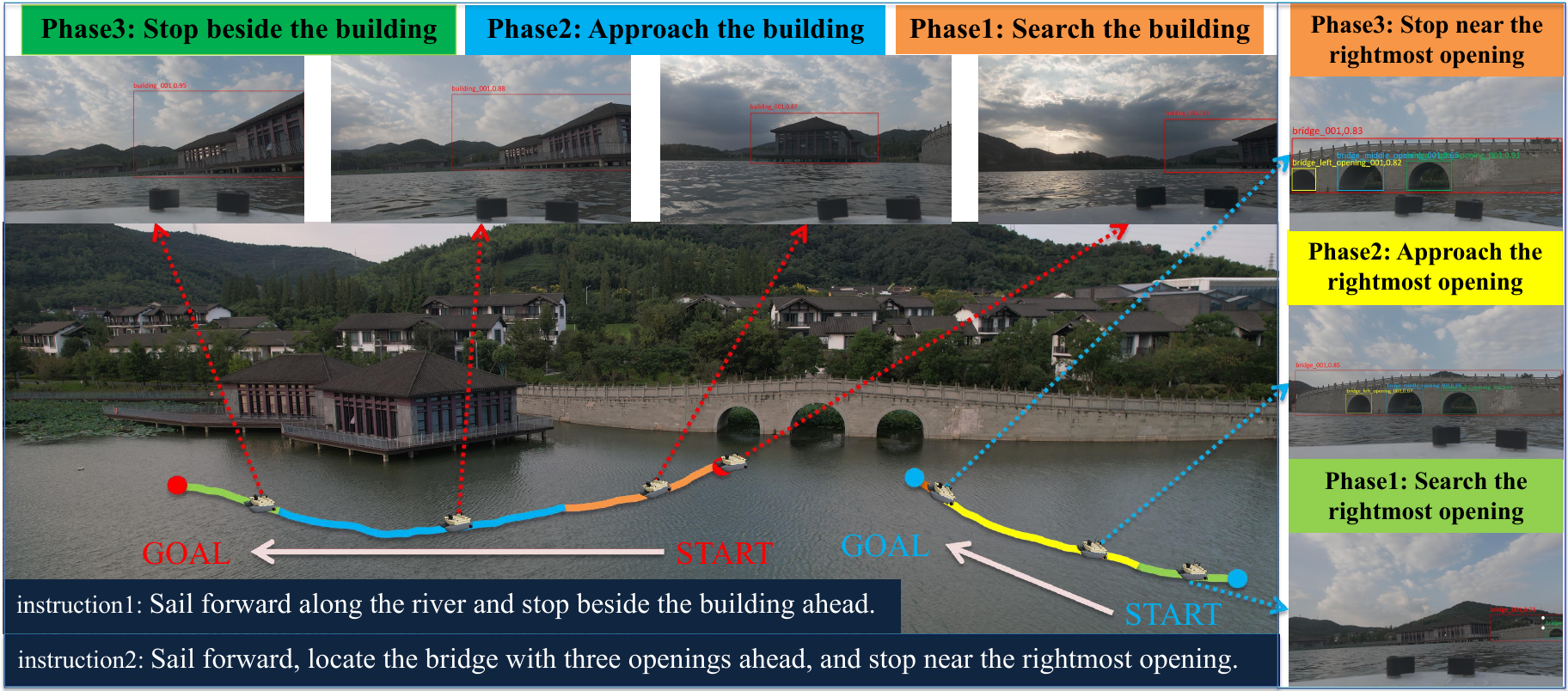}
    \caption{\textbf{Real-world closed-loop navigation of \policy{}.} The global trajectories and phase-aligned first-person observations show two representative tasks: stopping beside a building and stopping near the rightmost opening of a three-opening bridge.}
    \label{fig:real}
\end{figure*}

\subsection{Closed-Loop Navigation in Unity}

\begin{table}[!htbp]
\centering
\caption{Unity--ROS closed-loop performance}
\label{tab:unity}
\small
\setlength{\tabcolsep}{2.8pt}
\renewcommand{\arraystretch}{0.96}
\begin{NiceTabular}{lccccccc}
\toprule
\textbf{Task} & \textbf{SR} & \textbf{SCR} & \textbf{PLR} &
\textbf{Coll.} & \textbf{WPR} & \textbf{G-Acc.} & \textbf{P-Acc.} \\
\midrule
Long   & 0.60 & 0.745 & 1.03 & 0.40 & 0.10 & 0.945 & 0.65 \\
Bridge & 0.95 & 0.970 & 0.99 & 0.00 & 0.05 & 0.905 & 0.95 \\
Left   & 0.85 & 0.960 & 1.06 & 0.15 & 0.05 & 0.980 & 0.95 \\
Right  & 0.90 & 0.980 & 1.00 & 0.05 & 0.05 & 0.980 & 0.95 \\
L-Turn & 0.70 & 0.925 & 0.84 & 0.05 & 0.25 & 0.990 & 0.78 \\
R-Turn & 0.75 & 0.938 & 0.97 & 0.10 & 0.15 & 0.990 & 0.81 \\
\midrule
\textbf{Avg.}
& \textbf{0.79} & \textbf{0.920} & \textbf{0.98} & \textbf{0.13} &
\textbf{0.11} & \textbf{0.965} & \textbf{0.85} \\
\midrule
Test-Unseen & 0.80 & 0.913 & 0.95 & 0.10 & 0.05 & 0.95 & 0.95 \\
\bottomrule
\end{NiceTabular}
\par\vspace{2mm}
{\footnotesize
\raggedright
\noindent Note: Each task is evaluated over 20 trials.
\par}
\end{table}

In closed-loop simulation, each predicted local trajectory is executed before the next camera observation is acquired and the phase and grounding states are updated. We report success rate (SR), stage completion rate (SCR), path-length ratio (PLR), collision rate (Coll.), wrong-passage rate (WPR), grounding accuracy (G-Acc.), and phase accuracy (P-Acc.). A trial is successful only when all instruction-critical stages and the terminal condition are satisfied without collision; SCR measures the fraction of completed semantic stages, and PLR is defined as $L_{\mathrm{exec}}/L_{\mathrm{ref}}$, so a value below one may also indicate early termination. WPR records entry into an incorrect bridge opening, gate, or instruction-inconsistent direction.

\begin{figure}[!htbp]
    \centering
    \includegraphics[width=0.30\textwidth]{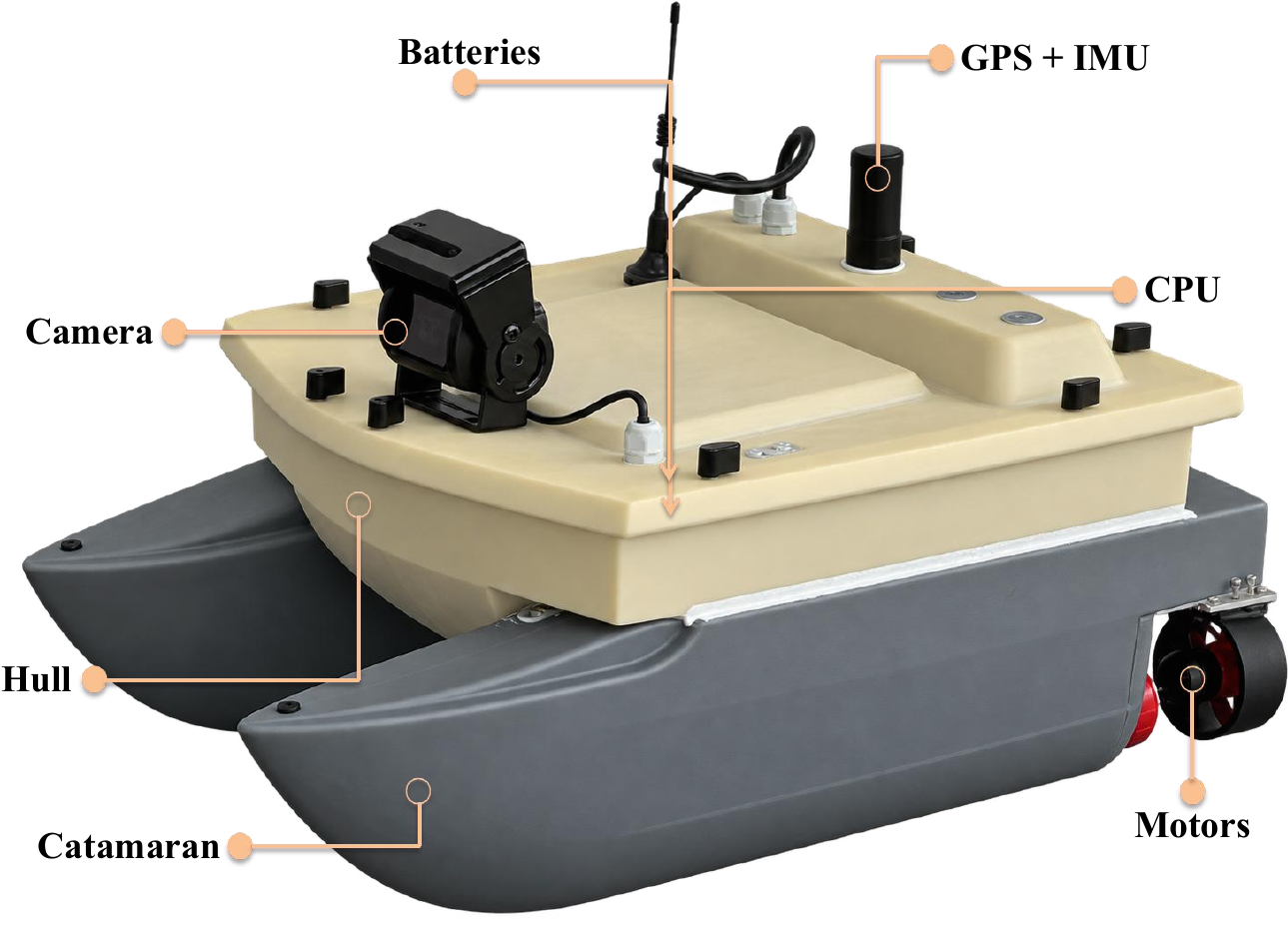}
    \caption{Physical USV platform developed for real-world \policy{} evaluation, integrating an RGB camera, GPS/IMU, onboard CPU, battery system, and dual motors.}
    \label{fig:real_platform}
\vspace{-4mm}
\end{figure}

Table~\ref{tab:unity} gives an average SR of 0.79 and SCR of 0.920 across the six closed-loop task categories. Bridge and left/right-opening tasks achieve 0.85--0.95 SR, whereas turning tasks exhibit higher WPR. Long-distance navigation is the most difficult category, with an SR of 0.60. These routes average 206.1\,m in length and contain 6.95 phases, exposing the system to more phase transitions and accumulated motion error, although SCR and G-Acc. remain 0.745 and 0.945, respectively. For the unseen closed-loop study, we conduct 20 trials on held-out left/right bridge-opening scenarios using manually specified navigation instructions. These trials form a closed-loop evaluation subset of Test-Unseen, which contains 23 routes with 6.35 phases per route on average. On this evaluation, \policy{} reaches 0.80 SR and 0.913 SCR, with both grounding and phase accuracy at 0.95; collision and WPR remain low at 0.10 and 0.05, respectively. Together, the results suggest that maintaining reliable phase transitions and motion execution over long routes remains more challenging than completing individual bridge-opening tasks.

\begin{table}[!htbp]
\centering
\caption{Closed-loop comparison and ablation on left/right bridge-opening tasks}
\label{tab:closed_loop_baseline}
\small
\setlength{\tabcolsep}{4.2pt}
\renewcommand{\arraystretch}{0.98}
\begin{NiceTabular}{lccccc}
\toprule
\textbf{Method} &
\textbf{SR$\uparrow$} &
\textbf{SCR$\uparrow$} &
\textbf{PLR} &
\textbf{Coll.$\downarrow$} &
\textbf{WPR$\downarrow$} \\
\midrule
GNM-style     & 0.25 & --   & 0.88 & 0.70 & 0.15 \\
ViNT-style    & 0.10 & --   & 0.97 & 0.90 & \textbf{0.00} \\
w/o Phase     & 0.00 & --   & --   & 1.00 & 0.30 \\
w/o Grounding & 0.15 & 0.25 & 1.21 & 0.85 & 0.40 \\
\midrule
\textbf{PGT-NAV} & \textbf{0.875} & \textbf{0.97} & 1.03
              & \textbf{0.10} & 0.05 \\
\bottomrule
\end{NiceTabular}
\end{table}

Table~\ref{tab:closed_loop_baseline} compares adapted visual-navigation baselines and key ablations on left/right bridge-opening tasks. ``--'' denotes an unavailable metric: SCR is omitted for methods without explicit phase-state tracking, and PLR for w/o Phase because no trial succeeds. \policy{} achieves 0.875 SR with a 0.10 collision rate, compared with 0.25/0.70 for GNM-style and 0.10/0.90 for ViNT-style. ViNT-style's WPR of 0.00 is not meaningful because most trials end in collision before a wrong passage is recorded. Removing the phase state yields no successful trials, while removing grounding reduces SR/SCR to 0.15/0.25 and raises WPR to 0.40. Together with the offline ablation, these results show that explicit phase progress and phase-target association improve language-conditioned navigation reliability.

\subsection{Real-World USV Experiments}

To evaluate deployment beyond simulation, we built the physical catamaran shown in Fig.~\ref{fig:real_platform}, integrating a forward RGB camera, GPS/IMU, onboard CPU, battery system, and dual motors. The onboard stack retains the same phase definitions and predict-six/execute-three interface used in simulation and runs perception, online phase update, trajectory prediction, and ROS communication without task-specific changes to the high-level policy.

We evaluate two language-conditioned tasks: stopping beside a specified building and locating a three-opening bridge before stopping near its rightmost opening. Figure~\ref{fig:real} shows the executed routes and phase-aligned observations. In both cases, the system progresses from target search to approach and terminal stopping despite reflections, sensor noise, vessel inertia, and control deviations, providing system-level evidence that the semantic phase representation and local trajectory interface can operate on the physical USV without task-specific scripting.

\subsection{Discussion}

Overall, the experiments indicate that the main benefit of \policy{} lies in maintaining semantic and motion consistency across repeated decisions rather than merely improving single-step waypoint prediction. The gap between teacher-forced and recursive evaluation shows that low local prediction error does not necessarily prevent accumulated position and heading drift. The ablation results further suggest that phase tracking, temporal context, and target grounding make complementary contributions to stable trajectory prediction. Closed-loop evaluation confirms that these improvements translate into higher navigation success and lower collision rates than the adapted visual-navigation baselines. The relatively lower performance on long multi-stage routes also highlights the importance of maintaining semantic progress and motion accuracy over successive replanning cycles.

The real-USV experiments further show that the phase-grounded representation and receding-horizon execution strategy can be transferred from simulation to the physical platform without task-specific changes to the high-level policy. Across the evaluated tasks, the system maintains semantic progress and completes language-conditioned navigation despite reflections, sensing noise, vessel inertia, and imperfect tracking. These results provide system-level evidence that the proposed framework supports continuous-motion navigation on a physical USV.
\section{CONCLUSION}

We presented \dataset{} and \policy{} for long-horizon vision--language navigation of unmanned surface vehicles in continuous river environments. Rather than mapping the full instruction directly to motion, \policy{} decomposes instructions into ordered semantic phases and updates the active phase using visual and motion evidence. By integrating semantic progress, temporal visual--motion context, and phase-specific grounding, the framework predicts short-horizon $\mathrm{SE}(2)$ waypoints and executes them through a receding-horizon closed-loop strategy. Experiments demonstrate that \policy{} reduces recursive trajectory and heading drift and achieves higher closed-loop navigation success than adapted visual-navigation baselines. Evaluation on unseen bridge-opening scenarios further supports its ability to execute language-conditioned navigation beyond the training routes. Real-world USV trials show that the phase-grounded navigation framework can operate on a physical USV without task-specific changes to the policy, demonstrating its applicability to continuous-motion river navigation.

\bibliographystyle{IEEEtran}
\bibliography{reference}

\end{document}